\documentclass[10pt,twocolumn,letterpaper]{article}
\usepackage[pagenumbers]{wacv}

\usepackage{microtype}
\usepackage{longtable}
\usepackage{array}
\usepackage{seqsplit}
\newcommand{\ev}[1]{\textsuperscript{\textsf{[#1]}}}

\definecolor{wacvblue}{rgb}{0.21,0.49,0.74}
\usepackage[breaklinks,colorlinks,allcolors=wacvblue]{hyperref}

\title{Fewer Tokens, More Answers:\\Verifiable-Reward Tuning of an Ornith-Style VLM\\for Multi-View Driving QA}

\author{T N Rickesh$^1$ \qquad Pranav Vinodh$^2$ \qquad Sowmya Kamath S.$^2$ \\
$^1$natkal \qquad $^2$National Institute of Technology Karnataka \\
{\tt\small info@natkal.ai \qquad prnv.231it050@nitk.edu.in \qquad sowmyakamath@nitk.edu.in}
}

\begin{document}
\maketitle

\begin{abstract}
Reasoning-enabled model derivatives must turn intermediate deliberation into a usable answer within a deployment budget. We study that transition in an Ornith checkpoint adapted to multi-camera driving question answering with verifiable rewards. Paired archived outputs show that, on a selected 400-question development screen with a 256-token cap, GRPO increases answer extraction from 78.25\% to 90.25\% and accuracy from 55.50\% to 63.75\%. Of 39 wrong-to-correct transitions, 33 begin with an unextractable base response. The base's 87 unextractable responses all reach the token cap. Checkpoint-verified reconstruction on those same questions gives a much smaller first-pass accuracy gain at 512 tokens: +0.75 percentage points, versus +8.25 at 256. A later answer-cued continuation is analyzed separately; it helps the base more and narrows the adapter difference. On the full 2,511-question development set, this recovery-enabled pipeline reduces mean generated tokens from 108.6 to 47.1, while the paired accuracy interval includes zero gain. We frame these findings as adaptation of answer delivery under finite budgets, with explicit selection and provenance limits. Ornith's self-scaffolding lineage and newer reasoning-enabled derivatives such as Bonsai motivate the problem, but our measured results concern Ornith alone and do not establish a shared architectural mechanism or cross-model transfer.
\end{abstract}

\section{Introduction}
A reasoning-capable vision-language model can use its response budget to deliberate without producing an answer that an application can extract. In a driving multiple-choice task, this creates an operational failure even if the partial response contains plausible scene analysis. In autonomous driving where real-time decisions face strict latency deadlines, this deliberation tax illustrates the operational fragility of verbose text-based reasoning under finite budgets, providing compelling motivation for latent visual representations that deliberate without verbose token emissions. The relevant deployment question is therefore not only whether post-training improves the final label, but whether the model reliably delivers that label before its generation budget is exhausted.

This problem matters for multimodal derivatives designed for reasoning, agentic use, or efficient local deployment. Their inherited visual backbone, post-training recipe, and output conventions may differ from the general-purpose checkpoints from which they originate. Ornith and the vision-enabled Bonsai models are examples of this broader ecosystem~\citep{ornithreport,bonsai}. The shared question concerns answer delivery; it does not require their compression methods or self-improvement procedures to be identical.

We examine an Ornith checkpoint on WaymoQA~\citep{waymoqa}, using verifiable-reward tuning to encourage a correct, extractable answer. Its large selected-screen gain under a shorter generation cap is mostly associated with recovered answer availability. A longer cap and an additional answer-cued continuation change that advantage. This makes the budget-dependent result a finding about an answer-collection bottleneck, rather than a caveat appended to a single accuracy score.

The empirical contribution is a paired protocol analysis: verify checkpoint identity, preserve every requested question in the denominator, distinguish initial generation from generative recovery, and relate scored improvements to the base extraction state. We report completion length as emitted tokens, without converting it to an unmeasured compute or latency claim. The resulting study is of answer delivery in one tested reasoning-enabled derivative, with a concrete protocol for examining others.

\paragraph{Companion relationship.}
The Gemma-family companion analyzes extracted-answer correction and residual category errors~\citep{companiongemma}. The two papers share a WaymoQA board and a small reference ladder, explicitly acknowledged here. Ornith's development outputs, budget comparison, and recovery analysis are the primary evidence in this manuscript. We do not reuse the companion's Gemma tuning result as another contribution or present two independent benchmark collections.

\section{Reasoning-Oriented Derivatives and Related Work}

\subsection{A reasoning-oriented derivative ecosystem}
``Ornith-style'' is our study label for the reasoning-enabled derivative setting exemplified by the tested checkpoint. It is not an official architecture taxonomy. We track three separate dimensions: backbone design, additional training or compression, and the reasoning/output mode used by the evaluator. Dense versus mixture-of-experts routing alone does not determine the presence or usefulness of intermediate deliberation.

Ornith documents a self-scaffolding training framework that learns task-specific orchestration together with solution rollouts, building on Gemma and Qwen backbones~\citep{ornithreport}. Its later generation extends the self-improvement loop~\citep{ornithlatest}. These descriptions characterize lineage and training, not proof that our driving responses execute a specific learned scaffold at inference. The historical local model directory does not authenticate a current branded release, so we identify the evaluated artifact by its recorded content hash.

Bonsai's documented vision-enabled models retain thinking and agentic behavior while using low-bit representations~\citep{bonsai,bonsaidemo}. This makes Bonsai relevant to the same finite-budget answer-delivery question. It does not establish Ornith's self-scaffolding procedure for Bonsai. Nor should a standalone compression scheme be equated with mixture-of-experts routing. Table~\ref{tab:ecosystem} separates these properties so that the broader model discussion does not become an unsupported architectural equivalence.

\begin{table}[t]
\centering\small\resizebox{\columnwidth}{!}{%
\begin{tabular}{p{.22\linewidth}p{.36\linewidth}p{.42\linewidth}}
\toprule
Model lineage & Documented distinction & Role in this paper \\
\midrule
Ornith & Learned scaffolds and self-improvement & Recorded checkpoint is experimentally studied; release identity remains incomplete \\
Bonsai & Compressed multimodal backbone with thinking support & Related derivative; no WaymoQA evaluation or tuning result reported \\
Gemma & Multiple native multimodal designs; configurable thinking & Shared diagnostic reference, with tuning studied in the companion \\
\bottomrule
\end{tabular}%
}
\caption{Primary-documentation context~\citep{ornithreport,ornithlatest,bonsai,gemmadocs}. Properties can overlap; these are not mutually exclusive architecture classes. External release descriptions are not our driving measurements.}
\label{tab:ecosystem}
\end{table}

\subsection{Test-Time Reasoning and Token Budgets in Vision-Language Models}
Recent post-training approaches incentivize models to generate multi-step deliberative reasoning chains before emitting an answer. In language tasks, methods such as DeepSeek-R1~\citep{deepseekr1} demonstrate that reinforcement learning with verifiable rewards encourages self-correction. Snell et al.~\citep{snell2024scaling} show that scaling test-time compute can match or exceed gains from scaling model parameter count. In visual domains, frameworks such as Visual-CoT~\citep{shao2024visualcot} and LLaVA-CoT~\citep{xu2024llavacot} train models to explicitly verbalize intermediate spatial relations, bounding boxes, and object attributes.

However, unconstrained verbal deliberation introduces practical failure modes. Chen et al.~\citep{chen2024overthinking} document the overthinking phenomenon, where models generate redundant intermediate text that adds latency without improving accuracy. In embodied and automotive domains, where vision-language models must process multiple high-resolution cameras under strict real-time deadlines, every emitted reasoning token incurs latency and inference cost. A critical failure mode under fixed generation caps is token exhaustion, where the model consumes its entire token allowance on preliminary narration and terminates before outputting the required prediction.

\subsection{Latent Visual Reasoning versus Explicit Verbalization}
The limitations of text-based deliberation have prompted research into whether spatial and physical reasoning belongs in explicit language or continuous latent spaces. LeCun~\citep{lecun2022path} argues that auto-regressive language generation is ill-suited for physical commonsense and continuous dynamics, proposing Joint Embedding Predictive Architectures (JEPA) that simulate future world states directly in latent representations.

Recent language model architectures explore hybrid approaches. Hao et al.~\citep{hao2024coconut} introduce Coconut, which replaces discrete reasoning tokens with continuous hidden-state transitions, allowing models to deliberate without generating language. In multimodal domains, Zhu et al.~\citep{zhu2024leveraging} demonstrate that models can leverage latent visual reasoning effectively without verbose textual decoding. When explicit linguistic deliberation is evaluated under a strict token cap, verifiable rewards can train the policy to compress its verbalization, routing reasoning through its latent representations rather than producing verbose textual chains.

\subsection{Vision-Language Models for Multi-View Driving Reasoning}
Autonomous driving perception requires reasoning across multiple camera views. DriveLM~\citep{drivelm} introduces graph-structured driving question answering connecting perception, prediction, and planning objects. WaymoQA~\citep{waymoqa} provides multi-view benchmark questions derived from real-world driving sequences across eight surrounding cameras.

Integrating vision-language models into driving pipelines requires balancing reasoning depth with real-time constraints. Tian et al.~\citep{tian2024drivevlm} introduce DriveVLM, which deploys a dual-system architecture combining a fast reactive planner with a slow deliberative visual reasoner. Chu et al.~\citep{chu2023mobilevlm} optimize vision-language backbones for mobile and edge compute, while FastV~\citep{chen2024fastv} accelerates inference through visual token pruning. Our work studies this boundary: rather than modifying visual architectures or deploying separate fast/slow modules, we measure how verifiable-reward post-training alters generation length and answer availability under finite token allowances.

\section{Data, Training, and Answer-Collection Protocols}
\subsection{Development membership and model selection}
The primary Ornith evidence contains paired predictions for a 400-question development screen and a 2,511-question development set.\ev{E5--E6} Evaluation artifacts content-hash the selected question files, model directory, adapter directory, benchmark manifest, and frame inventory. We verify screen membership through benchmark identifiers and matching scene, frame, camera views, category, and label.

The screen is a question subset of development, and both cover the same 30 scene tokens. All of those scenes also appear outside the screen.\ev{E5--E6} Therefore the remaining development questions do not form an independent scene holdout. The screen was used to compare candidate learning rates; its largest reported gain is post-selection. Bootstrap intervals and item-level tests are conditional descriptions of these archived data, not selection-adjusted confirmation.

The training workflow is pinned to a manifest-based partitioning implementation, but the complete historical training manifest is not recovered.\ev{E7} Code intent and artifact-directory names are insufficient to certify actual training/development scene disjointness. We retain that unresolved boundary, and do not relabel the larger development evaluation as test.

\subsection{Recorded verifiable-reward recipe}
The workflow-pinned profile specifies QLoRA rank 16 and scaling 32 on q/k/v/o attention projections, four sampled completions, temperature 1.0, a 256-token training completion cap, 300-step candidates, and learning rates $\{3\!\times\!10^{-7},10^{-6},3\!\times\!10^{-6}\}$.\ev{E7,E11--E12,E5} The profile sets a zero KL coefficient and optimizes a composite verifiable reward $R(y, y^*) = R_{\text{correct}}(y, y^*) + R_{\text{format}}(y) + R_{\text{extract}}(y)$, where $y$ is the generated output, $y^*$ is the ground-truth target, and $\mathcal{E}(y)$ denotes the extracted option:
\begin{equation}
R_{\text{correct}}(y, y^*) = \begin{cases} 
+2.0, & \text{if } \mathcal{E}(y) = y^* \\ 
-1.0, & \text{if } \mathcal{E}(y) \neq y^* \land \mathcal{E}(y) \neq \emptyset \\ 
-2.0, & \text{if } \mathcal{E}(y) = \emptyset,
\end{cases}
\end{equation}
with strict JSON format reward $R_{\text{format}}(y) = \pm 0.5$ and extractability reward $R_{\text{extract}}(y) = \pm 0.25$.

The selected adapter is the candidate at $3\!\times\!10^{-6}$. The profile values describe the workflow lineage; complete child-trainer state and historical split artifacts remain missing. We make no reward-component causal claim, no monotonic learning-rate dose-response claim, and no claim of advantage over a matched supervised answer-only adaptation.

\subsection{Single pass versus answer recovery}
The deterministic evaluator uses NF4 quantization, greedy decoding, one question per request, and aspect-preserving image resizing to a maximum dimension of 1,024 pixels.\ev{E5--E6} Each input supplies eight surrounding-camera views in the task-recorded order and asks for one JSON answer. A literal full-response JSON check measures strict compliance. The fallback extractor accepts an unambiguous embedded answer JSON or a short answer letter.

At a 256-token cap, the archived screen has only an initial generation.\ev{E5} The later 512-token evaluator can request up to 16 additional tokens when the initial completion has no extractable answer. It appends ``Therefore, the answer is'' to the existing assistant text, generates a continuation, and attempts extraction from that tail.\ev{E6} Formally, for input $x$ under generation budget $B$, the two-stage recovery policy generates:
\begin{equation}
\hat{y} = \begin{cases}
\mathcal{E}(f_\theta(x; B)), & \text{if } \mathcal{E}(f_\theta(x; B)) \neq \emptyset \\
\mathcal{E}\big(f_\theta([x; f_\theta(x; B); c_{\text{cue}}]; B_{\text{rec}})\big), & \text{otherwise},
\end{cases}
\end{equation}
where $c_{\text{cue}}$ appends ``Therefore, the answer is'' and $B_{\text{rec}} = 16$. This \emph{recovery} is a model call and can change the answer; it is not a passive parser improvement. The resulting total generation budget can exceed the initial cap.

For every continuation-marked response, we remove the appended cue and tail to reconstruct first-pass scoring, then verify that the original prefix has no extractable answer under the frozen parser. Noncontinued responses must reproduce their archived predictions. Matching model and adapter content hashes establish checkpoint identity between the 256- and 512-token tasks. The earlier 128-token task has a different adapter hash despite its matching directory name, and is excluded from a same-checkpoint budget curve.\ev{E10}

\subsection{Measures and statistical interpretation}
We keep all requested questions in the accuracy denominator, scoring missing or unextractable responses as wrong. Extraction rate measures the presence of an option label; strict rate measures compliance with the JSON-only contract. Neither is a causal grounding metric. Joint states cross correctness with strict extraction, fallback extraction, or unavailability. For wrong-to-correct pairs we record whether the base answer was already extractable.

Scene-cluster bootstrap intervals use 10,000 resamples with seed 42, drawing scene tokens with replacement and computing pooled item-weighted paired differences. The intervals address within-scene question dependence but not candidate selection, training-seed variation, or adaptive protocol design. Exact item-level McNemar results are secondary. Token counts are measured generation output; without timing, hardware, and prompt-processing measurements they are not FLOPs, energy, or end-to-end latency estimates.

\section{Diagnostic Motivation and Answer Availability}
\begin{table}[t]
\centering\small\resizebox{\columnwidth}{!}{\begin{tabular}{llrr}
\toprule
Checkpoint & Role & Correct / all & Acc. all (\%) \\
\midrule
Gemma-4-12B & Zero-shot reference & 4,072 / 5,519 & 73.78 \ev{E1} \\
Gemma-4-31B & Zero-shot reference & 4,261 / 5,519 & 77.21 \ev{E13} \\
Ornith + GRPO & Deployment observation & 3,812 / 5,519 & 69.07 \ev{E16} \\
\bottomrule
\end{tabular}
}
\caption{Shared-board reference context.\ev{E1,E13,E16} Services and generation settings differ. Ornith is a tuned deployment observation without a corresponding base run on this board. These rows are not a paired Ornith effect or a controlled architecture ranking.}
\label{tab:board}
\end{table}

The shared board contains 5,519 questions.\ev{E4} It supplies a common diagnostic reference: zero-shot Gemma checkpoints already yield recoverable answers and substantial accuracy, while the recorded Ornith adapter deployment returns a label on all board items at 69.07\% accuracy.\ev{E1,E13,E16} Because no matching Ornith base evaluation exists on this board, Table~\ref{tab:board} cannot demonstrate adaptation gain. The deployment task also suppresses thinking and requests JSON, so it does not authenticate a scaffold-preserving inference mode.

The paired development screen provides the relevant diagnostic. Before tuning, only 313 of 400 answers are extractable, and 222 are correct.\ev{E5} All 87 unextractable base responses use the complete 256-token allowance. The adapter has 39 unextractable responses, of which 38 reach that cap. These are observed token-budget events; no claim is made that a correct latent answer was already formed inside a truncated response.

\begin{figure}[t]
\centering\includegraphics[width=\linewidth]{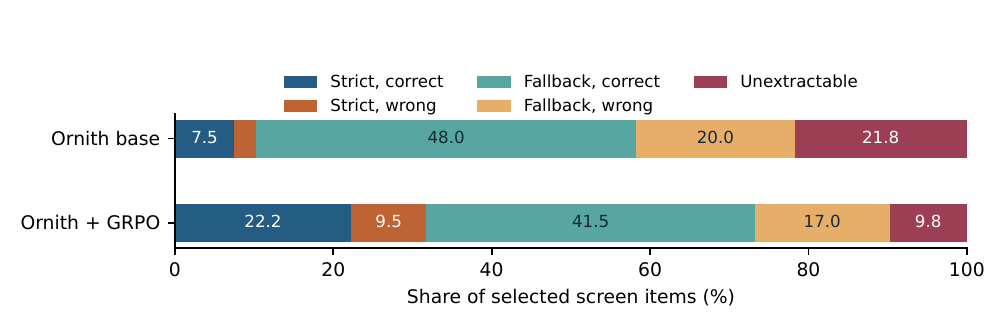}
\caption{Correctness and answer availability on the selected 256-token screen.\ev{E5} Correct fallback responses remain correct. Unextractable responses count against accuracy, but are not assigned a visual failure cause. This diagnostic differs from the companion's almost fully extractable Gemma base.}
\label{fig:availability}
\end{figure}

\section{Results: Concise Delivery under Finite Budgets}
\subsection{The Deliberation Tax: Token Exhaustion under Finite Budgets}
Table~\ref{tab:paired} and Figure~\ref{fig:availability} report results on the 400-question development screen under a 256-token cap. On this benchmark, the base Ornith model achieves 55.50\% accuracy (222/400) and extracts answers for only 78.25\% of questions (313/400).\ev{E5}

An audit of base model failures reveals that the principal failure mode is not inaccurate visual perception, but token budget exhaustion. All 87 unextractable base responses reach the exact 256-token cap. Instead of emitting the requested answer format, the base model generates intermediate verbal narration describing individual camera frames until the generation window terminates mid-sentence.

Following verifiable-reward post-training, accuracy increases to 63.75\% (255/400, +8.25 percentage points, scene CI [+5.23, +11.14], McNemar $p=5.42\times10^{-7}$), and extraction increases to 90.25\% (361/400).\ev{E5} As illustrated in Figure~\ref{fig:availability}, out of 39 gross improvements, 33 begin from an unextractable base response, whereas only 6 represent corrections of an extractable wrong answer. The adapter reduces mean generation length from 79.3 to 36.4 tokens. These measurements show that under short budgets, the observed gain is driven by resolving an answer-collection bottleneck: the model learns to compress its verbalization so that the answer is delivered within the assigned budget.

\subsection{The Budget Crossover Effect: Disentangling Concision from Reasoning Capacity}
To determine whether GRPO elevated the model's fundamental visual reasoning capacity or simply accelerated answer emission, we evaluate both models under a relaxed 512-token cap across the exact same 400 screen questions.

When given 512 tokens for first-pass generation, base model accuracy rises from 55.50\% to 68.50\% (274/400) purely because the model has room to complete its verbal reasoning before emitting the answer (Table~\ref{tab:paired}).\ev{E6} Adapter accuracy reaches 69.25\% (277/400). Consequently, expanding the token budget narrows the adapter advantage from +8.25 percentage points to +0.75 percentage points (scene CI [-2.23, +3.80]).

When answer-cued recovery is enabled (appending ``Therefore, the answer is'' to unextractable prefixes), both the base model and the adapter achieve identical scores: 279 correct answers out of 400 (69.75\%, $\Delta = 0.00$ pp, scene CI [-2.77, +3.09]).\ev{E6}

This interaction, visualized in Figure~\ref{fig:protocols}, isolates the mechanism of improvement. Because the base model reaches the adapter's performance ceiling once granted sufficient generation budget or an explicit prompt intervention, the post-training reward did not improve cross-camera spatial reasoning or feature representations. Instead, it operated strictly on answer delivery timing.

\begin{table}[t]
\centering\small\resizebox{\columnwidth}{!}{\begin{tabular}{lrrrrl}
\toprule
Data / protocol & $n$ & Base & GRPO & $\Delta$ (pp) & Scene CI (pp) \\
\midrule
Screen / 256 first pass & 400 & 55.50 & 63.75 & +8.25 & [+5.23, +11.14] \\
Screen / 512 first pass & 400 & 68.50 & 69.25 & +0.75 & [-2.23, +3.80] \\
Screen / 512 + recovery & 400 & 69.75 & 69.75 & +0.00 & [-2.77, +3.09] \\
Dev / 512 first pass & 2,511 & 68.66 & 70.45 & +1.79 & [+0.45, +3.23] \\
Dev / 512 + recovery & 2,511 & 70.53 & 71.68 & +1.15 & [-0.12, +2.42] \\
\bottomrule
\end{tabular}
}
\caption{Paired Ornith accuracy (\%) by data and collection protocol.\ev{E5--E6} Intervals are scene-cluster summaries conditional on the selected configuration. Screen questions are inside development; these rows are not independent evaluations.}
\label{tab:paired}
\end{table}

\subsection{Post-Training as a Test-Time Compute Compression Operator}
On all 2,511 development questions, evaluating first-pass generation at 512 tokens yields 68.66\% accuracy for the base model (1,724/2,511) and 70.45\% for the adapter (1,769/2,511), a net gain of +1.79 percentage points (scene CI [+0.45, +3.23]).\ev{E6} Enabling generative recovery adds 47 correct answers to the base model and 31 to the adapter, resulting in recovery-enabled accuracies of 70.53\% (1,771/2,511) and 71.68\% (1,800/2,511, $\Delta = +1.15$ pp, scene CI [-0.12, +2.42], item-level McNemar $p=0.055$).

\begin{figure}[t]
\centering\includegraphics[width=\linewidth]{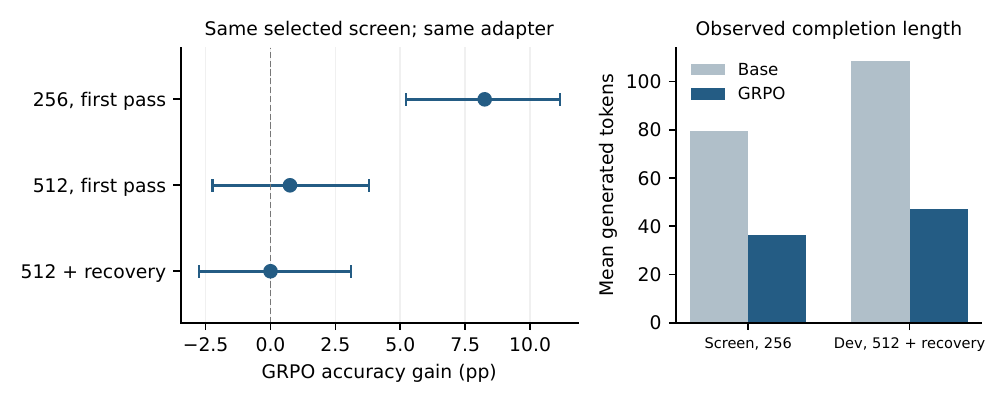}
\caption{Same-screen accuracy gains under distinct collection protocols, and recorded completion lengths.\ev{E5--E6} The cost panel compares a single-pass screen protocol and the larger recovery-enabled development protocol; it does not estimate equal-workload latency.}
\label{fig:protocols}
\end{figure}

While the paired accuracy difference on the full development set is modest, the generation length difference is substantial. As shown in the cost panel of Figure~\ref{fig:protocols}, mean generated tokens across the full development set fall from 108.6 tokens for the base pipeline to 47.1 tokens for the adapter pipeline, a 56.6\% reduction in emitted length.\ev{E6}

For real-time autonomous systems, this demonstrates that verifiable-reward post-training functions effectively as a test-time compression operator. The adapter preserves the accuracy achieved by the base model with verbose deliberation while cutting the generation footprint by more than half. In multi-camera systems, end-to-end latency is jointly governed by visual token prefill across eight cameras and auto-regressive token decoding. Compressing emitted tokens provides essential operational buffer against decoding timeouts in embedded vehicle computers, while visual token pruning remains complementary for the prefill stage.

\section{Implications for Ornith-Style and Other Reasoning Derivatives}
The results locate an adaptation target: finite-budget answer delivery. At the shorter cap, many scored improvements replace unavailable responses. Once the base receives a longer initial budget or a recovery cue, the selected adapter advantage narrows. Describing the finding through output availability and budget is more informative than treating each score as an invariant measure of reasoning competence.

This question extends to other reasoning-enabled derivatives, including vision-capable Bonsai models, because they also expose thinking modes~\citep{bonsai,bonsaidemo}. The present results do not establish that those models truncate similarly, benefit from this reward, or share Ornith's self-scaffolding training. Compression, backbone design, post-training, and output mode must each be recorded and tested. We include those models as related candidates, not as unperformed WaymoQA experiments.

An extension should fingerprint each checkpoint and image processor, verify that the evaluated release accepts the intended multi-image input, and compare first-pass budgets with recovery as a separate condition on untouched scenes. An eight-image request is not enough to prove that a backend actually consumed eight distinct views. Source-defined multimodal support and run-level evidence of image handling serve different purposes.

\section{Limitations and Conclusion}
The main limitations are candidate selection on the reported screen, screen/development scene reuse, incomplete historical training membership, a single tested Ornith checkpoint, and absence of matched SFT or reward ablations. The unpaired deployment board cannot supply the missing independent base/adapter effect. Category errors do not establish failed coordinate transformations, uncertainty calibration, or causal planning mechanisms. There is also no verified measurement of latency or inference compute.

Within those limits, verifiable-reward tuning changes how the tested checkpoint delivers an answer. The selected shorter-budget result combines improved availability and shorter completions; larger-budget and recovery analyses delimit the accuracy claim. These findings motivate a protocol-centered study of reasoning-enabled derivatives, while keeping Bonsai and later Ornith releases as documented context until task-linked measurements exist.

{
    \small
    \bibliographystyle{ieeenat_fullname}
    \bibliography{references_lvr}
}

\appendix
\section{Checkpoint Identity and Recovery Reconstruction}
Both included generation-budget tasks record adapter hash
\begin{flushleft}\small\texttt{\seqsplit{b014343d1db667c048b64e3360bedb91275c5be7f8a43ecd408988620fba57d0}}\end{flushleft}
and model hash
\begin{flushleft}\small\texttt{\seqsplit{94e6a9d3cdc7d10e0a197a26dd96e0e39b347b6fb30ec97b8cc7c5ba94270979}}.\end{flushleft}
The excluded 128-token run has adapter hash
\begin{flushleft}\small\texttt{\seqsplit{89b3c48dc78e53413617038d557d9e3bdbcb08234b41fb2e003e2f18ed595be1}}.\end{flushleft}
Directory names do not supply immutable model identity.\ev{E5--E6,E10}

The evaluator revisions are \texttt{522c2519} at 256 tokens and \texttt{9e587eed} for the later protocol. Their archived code fingerprints and pinned source identify the answer-recovery addition. For each marked response, reconstruction removes the last appended cue and continuation, checks that the prefix had no extracted answer, and re-scores with the frozen contract parser. All noncontinued records must agree with their archived predictions. First-pass tokens are not separately reconstructed; emitted-token statistics describe the full recorded pipeline.

The development JSONL and screen JSONL have recorded hashes and benchmark identifiers.\ev{E5--E6} Item-level comparisons confirm their subset relationship and their shared 30 scenes. The source workflow records a pinned training implementation, but the complete training manifest is absent from this cache. The derived intervals therefore retain both selection and partition-provenance qualifications.

\section{Learning-Rate Candidates}
\begin{table}[h]\centering\small\setlength{\tabcolsep}{3pt}\begin{tabular}{lrrr}
\toprule
Candidate learning rate & Correct & Gain (pp) & Source \\
\midrule
$3\times10^{-7}$ & 227 / 400 & +1.25 & E11 \\
$10^{-6}$ & 226 / 400 & +1.00 & E12 \\
$3\times10^{-6}$ & 255 / 400 & +8.25 & E5 \\
\bottomrule
\end{tabular}

\caption{Archived 256-token screen candidates.\ev{E5,E11--E12} Base accuracy is 222/400. The largest candidate was selected using this screen; its uncertainty is not a prospective confirmation or a general dose-response result.}\end{table}

\section{Shared Evidence and Reproduction}
Reference-board CSV membership is checked against the authenticated board's ordered scene, category, and labels. Gemma question strings, truncated in those artifacts, match prefixes of board questions.\ev{E1,E4,E13} The Ornith deployment task records thinking disabled, JSON forcing, and a 256-token cap.\ev{E16} The archive has no matching board base run, and the deployment artifact's adapter-directory lineage does not certify content identity with the paired development adapter.

All measured results in this manuscript come from the recorded evaluation tasks. Vendor model documentation supplies ecosystem context only; no external model score is presented as our WaymoQA result. The shared author cache and scripts reproduce paired statistics, recovery accounting, tables, and vector figures without new inference. The companion studies overlap in reference data, not in their primary tuning outcome. Any simultaneous submission must disclose that overlap according to the receiving venues' actual policies.

\end{document}